\documentclass[10pt,twocolumn,letterpaper]{article}

\usepackage{iccv}              

\usepackage{colortbl}
\usepackage{multirow}
\usepackage{algorithm}
\usepackage{algorithmic}
\usepackage{amsfonts}

\definecolor{iccvblue}{rgb}{0.21,0.49,0.74}
\usepackage[pagebackref,breaklinks,colorlinks,allcolors=iccvblue]{hyperref}

\def\paperID{*****} 
\def\confName{ICCV}
\def\confYear{2025}

\title{Open-Ended 3D Point Cloud Instance Segmentation} 

\author{ Phuc Nguyen$^{1,*}$\and
Minh Luu$^{1,*}$ \and
Anh Tran$^{1}$ \and
Cuong Pham$^{1,2}$ \and
Khoi Nguyen$^{1}$ \and
\normalsize{$^1$Movian AI \qquad $^2$Posts \& Telecommunications Inst. of Tech} \\
{\tt\small \{phucnda,  hoangminh291101, anstar1111, ducminhkhoi\}@gmail.com} \qquad \tt\small{cuongpv@ptit.edu.vn}}

\begin{document}
\def\mA{\mathcal{A}}
\def\mB{\mathcal{B}}
\def\mC{\mathcal{C}}
\def\mD{\mathcal{D}}
\def\mE{\mathcal{E}}
\def\mF{\mathcal{F}}
\def\mG{\mathcal{G}}
\def\mH{\mathcal{H}}
\def\mI{\mathcal{I}}
\def\mJ{\mathcal{J}}
\def\mK{\mathcal{K}}
\def\mL{\mathcal{L}}
\def\mM{\mathcal{M}}
\def\mN{\mathcal{N}}
\def\mO{\mathcal{O}}
\def\mP{\mathcal{P}}
\def\mQ{\mathcal{Q}}
\def\mR{\mathcal{R}}
\def\mS{\mathcal{S}}
\def\mT{\mathcal{T}}
\def\mU{\mathcal{U}}
\def\mV{\mathcal{V}}
\def\mW{\mathcal{W}}
\def\mX{\mathcal{X}}
\def\mY{\mathcal{Y}}
\def\mZ{\mathcal{Z}} 

\def\bbN{\mathbb{N}} 
\def\bbR{\mathbb{R}} 
\def\bbP{\mathbb{P}} 
\def\bbQ{\mathbb{Q}} 
\def\bbE{\mathbb{E}}

\def\1n{\mathbf{1}_n}
\def\0{\mathbf{0}}
\def\1{\mathbf{1}}

\def\A{{\bf A}}
\def\B{{\bf B}}
\def\C{{\bf C}}
\def\D{{\bf D}}
\def\E{{\bf E}}
\def\F{{\bf F}}
\def\G{{\bf G}}
\def\H{{\bf H}}
\def\I{{\bf I}}
\def\J{{\bf J}}
\def\K{{\bf K}}
\def\L{{\bf L}}
\def\M{{\bf M}}
\def\N{{\bf N}}
\def\O{{\bf O}}
\def\P{{\bf P}}
\def\Q{{\bf Q}}
\def\R{{\bf R}}
\def\S{{\bf S}}
\def\T{{\bf T}}
\def\U{{\bf U}}
\def\V{{\bf V}}
\def\W{{\bf W}}
\def\X{{\bf X}}
\def\Y{{\bf Y}}
\def\Z{{\bf Z}}

\def\a{{\bf a}}
\def\b{{\bf b}}
\def\c{{\bf c}}
\def\d{{\bf d}}
\def\e{{\bf e}}
\def\f{{\bf f}}
\def\g{{\bf g}}
\def\h{{\bf h}}
\def\i{{\bf i}}
\def\j{{\bf j}}
\def\k{{\bf k}}
\def\l{{\bf l}}
\def\m{{\bf m}}
\def\n{{\bf n}}
\def\o{{\bf o}}
\def\p{{\bf p}}
\def\q{{\bf q}}
\def\r{{\bf r}}
\def\s{{\bf s}}
\def\t{{\bf t}}
\def\u{{\bf u}}
\def\v{{\bf v}}
\def\w{{\bf w}}
\def\x{{\bf x}}
\def\y{{\bf y}}
\def\z{{\bf z}}

\def\balpha{\mbox{\boldmath{$\alpha$}}}
\def\bbeta{\mbox{\boldmath{$\beta$}}}
\def\bdelta{\mbox{\boldmath{$\delta$}}}
\def\bgamma{\mbox{\boldmath{$\gamma$}}}
\def\blambda{\mbox{\boldmath{$\lambda$}}}
\def\bsigma{\mbox{\boldmath{$\sigma$}}}
\def\btheta{\mbox{\boldmath{$\theta$}}}
\def\bomega{\mbox{\boldmath{$\omega$}}}
\def\bxi{\mbox{\boldmath{$\xi$}}}
\def\bnu{\mbox{\boldmath{$\nu$}}}                                  
\def\bphi{\mbox{\boldmath{$\phi$}}}
\def\bmu{\mbox{\boldmath{$\mu$}}}

\def\bDelta{\mbox{\boldmath{$\Delta$}}}
\def\bOmega{\mbox{\boldmath{$\Omega$}}}
\def\bPhi{\mbox{\boldmath{$\Phi$}}}
\def\bLambda{\mbox{\boldmath{$\Lambda$}}}
\def\bSigma{\mbox{\boldmath{$\Sigma$}}}
\def\bGamma{\mbox{\boldmath{$\Gamma$}}}
                                  
\newcommand{\myprob}[1]{\mathop{\mathbb{P}}_{#1}}

\newcommand{\myexp}[1]{\mathop{\mathbb{E}}_{#1}}

\newcommand{\mydelta}[1]{1_{#1}}

\newcommand{\myminimum}[1]{\mathop{\textrm{minimum}}_{#1}}
\newcommand{\mymaximum}[1]{\mathop{\textrm{maximum}}_{#1}}    
\newcommand{\mymin}[1]{\mathop{\textrm{minimize}}_{#1}}
\newcommand{\mymax}[1]{\mathop{\textrm{maximize}}_{#1}}
\newcommand{\mymins}[1]{\mathop{\textrm{min.}}_{#1}}
\newcommand{\mymaxs}[1]{\mathop{\textrm{max.}}_{#1}}  
\newcommand{\myargmin}[1]{\mathop{\textrm{argmin}}_{#1}} 
\newcommand{\myargmax}[1]{\mathop{\textrm{argmax}}_{#1}} 
\newcommand{\myst}{\textrm{s.t. }}

\newcommand{\denselist}{\itemsep -1pt}
\newcommand{\sparselist}{\itemsep 1pt}

\definecolor{pink}{rgb}{0.9,0.5,0.5}
\definecolor{purple}{rgb}{0.5, 0.4, 0.8}   
\definecolor{gray}{rgb}{0.3, 0.3, 0.3}
\definecolor{mygreen}{rgb}{0.2, 0.6, 0.2}

\newcommand{\cyan}[1]{\textcolor{cyan}{#1}}
\newcommand{\blue}[1]{\textcolor{blue}{#1}}
\newcommand{\magenta}[1]{\textcolor{magenta}{#1}}
\newcommand{\pink}[1]{\textcolor{pink}{#1}}
\newcommand{\green}[1]{\textcolor{green}{#1}} 
\newcommand{\gray}[1]{\textcolor{gray}{#1}}    
\newcommand{\mygreen}[1]{\textcolor{mygreen}{#1}}    
\newcommand{\purple}[1]{\textcolor{purple}{#1}}       

\definecolor{greena}{rgb}{0.4, 0.5, 0.1}
\newcommand{\greena}[1]{\textcolor{greena}{#1}}

\definecolor{bluea}{rgb}{0, 0.4, 0.6}
\newcommand{\bluea}[1]{\textcolor{bluea}{#1}}
\definecolor{reda}{rgb}{0.6, 0.2, 0.1}
\newcommand{\reda}[1]{\textcolor{reda}{#1}}

\def\changemargin#1#2{\list{}{\rightmargin#2\leftmargin#1}\item[]}
\let\endchangemargin=\endlist
                                               
\newcommand{\cm}[1]{}

\newcommand{\mhoai}[1]{{\color{magenta}\textbf{[MH: #1]}}}

\newcommand{\mtodo}[1]{{\color{red}$\blacksquare$\textbf{[TODO: #1]}}}
\newcommand{\myheading}[1]{\vspace{1ex}\noindent \textbf{#1}}
\newcommand{\htimesw}[2]{\mbox{$#1$$\times$$#2$}}


\newif\ifshowsolution
\showsolutiontrue

\ifshowsolution  
\newcommand{\Solution}[2]{\paragraph{\bf $\bigstar $ SOLUTION:} {\sf #2} }
\newcommand{\Mistake}[2]{\paragraph{\bf $\blacksquare$ COMMON MISTAKE #1:} {\sf #2} \bigskip}
\else
\newcommand{\Solution}[2]{\vspace{#1}}
\fi

\newcommand{\truefalse}{
\begin{enumerate}
	\item True
	\item False
\end{enumerate}
}

\newcommand{\yesno}{
\begin{enumerate}
	\item Yes
	\item No
\end{enumerate}
}

\newcommand{\Sref}[1]{Sec.~\ref{#1}}
\newcommand{\Eref}[1]{Eq.~(\ref{#1})}
\newcommand{\Fref}[1]{Fig.~\ref{#1}}
\newcommand{\Tref}[1]{Table~\ref{#1}}

\definecolor{gray}{rgb}{0.3, 0.3, 0.3}

\twocolumn[{
\renewcommand\twocolumn[1][]{#1}%
\maketitle

\vspace{-30pt}
\begin{center}%
\includegraphics[width=0.9\linewidth]{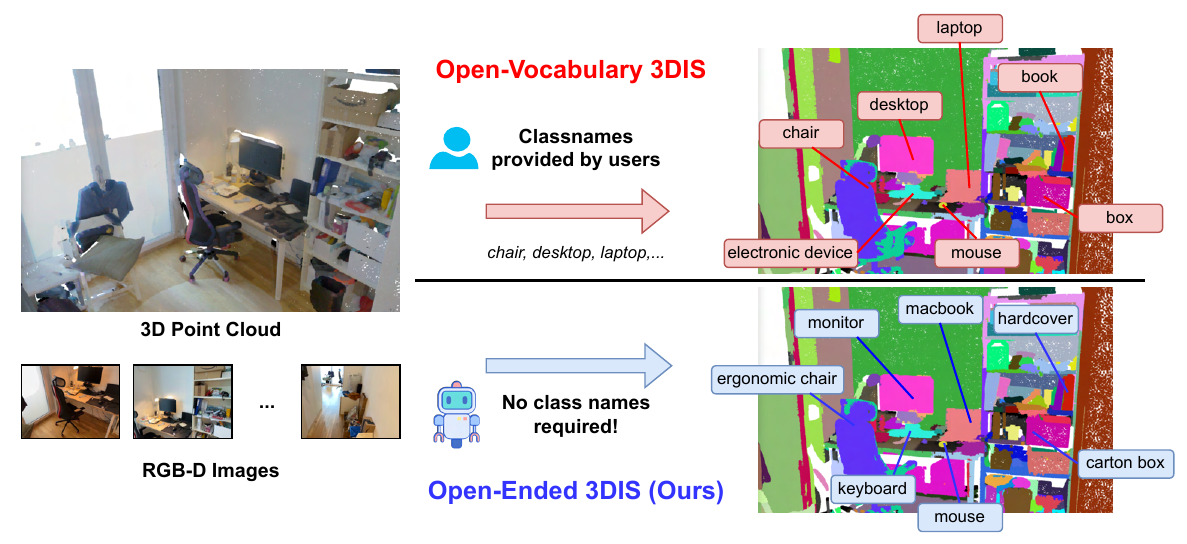}
\vspace{-5pt}
\captionof{figure}{During testing, Open-Vocabulary 3D Instance Segmentation (OV-3DIS) utilizes CLIP to generate confidence scores for provided text prompts. However, its reliance on a predefined vocabulary during inference necessitates human intervention, restricting autonomy in intelligent agents. In contrast, our proposed Open-Ended 3D Instance Segmentation (OE-3DIS) leverages Large Language Models to enable open interaction with humans, recognizing fine-grained objects without being constrained by predefined class labels.}
\label{fig:teaser}
\end{center}
}]
\def\thefootnote{*}\footnotetext{These authors contributed equally to this work}\def\thefootnote{\arabic{footnote}}

\begin{abstract}
\label{sec:abstract}

Open-vocabulary 3D Instance Segmentation methods (OV-3DIS) have recently demonstrated their generalization ability to unseen objects. However, these methods still depend on predefined class names during inference, restricting agents' autonomy. To mitigate this constraint, we propose a novel problem termed \textbf{Open-Ended 3D Instance Segmentation (OE-3DIS)}, which eliminates the necessity for predefined class names during testing. We present a comprehensive set of strong baselines inspired by OV-3DIS methodologies, utilizing 2D Multimodal Large Language Models. In addition, we introduce a novel token aggregation strategy that effectively fuses information from multiview images. To evaluate the performance of our OE-3DIS system, we benchmark both the proposed baselines and our method on two widely used indoor datasets: ScanNet200 and ScanNet++. Our approach achieves substantial performance gains over the baselines on both datasets. Notably, even without access to ground-truth object class names during inference, our method outperforms Open3DIS, the current state-of-the-art in OV-3DIS. Source code available at: \url{https://github.com/PhucNDA/OE-3DIS}.


\end{abstract}    
\section{Introduction}
\label{sec:introduction}
3D point cloud instance segmentation (3DIS) \cite{chen2021hierarchical, Schult23mask3d, ngo2023isbnet, he2021dyco3d, vu2022softgroup, jiang2020pointgroup}, also known as closed-vocabulary 3D instance segmentation, aims to segment all points in a point cloud into instances of classes predefined in the training set. However, this approach is less practical for scenarios where the test classes are unknown or different from the training classes. This limitation has led to the development of open-vocabulary 3D instance segmentation (OV-3DIS) \cite{nguyen2023open3dis, takmaz2023openmask3d, lu2023ovir, yin2023sai3d, yan2024maskclustering, Huang2023Segment3D}. Despite these advancements, OV-3DIS methods face several practical challenges. One major challenge is that class names must be predefined during inference, necessitating human intervention for scene understanding. This requirement significantly impedes the perception of truly autonomous agents. 
A potential solution is to predefined large and comprehensive vocabularies; however, this can lead to inaccuracies and significantly degrade the performance of OV-3DIS when an excessive number of vocabularies are used.

To overcome these limitations, we introduce a novel task called \textbf{Open-Ended 3D Point Cloud Instance Segmentation (OE-3DIS)} see \cref{fig:teaser}. Unlike traditional methods, OE-3DIS does not require predefined class names during testing. Given a 3D point cloud and RGBD sequence, the system automatically generates a set of 3D masks along with their class names. To evaluate the performance of OE-3DIS methods, we follow \cite{Rewatbowornwong2023ZeroGuideSeg} to leverage an adaptive protocol of standard AP score calculation.
To tackle this new and challenging task, \textbf{we introduce a novel method along with our proposed three established baselines} that leverage OV-3DIS techniques and Multimodal Large Language Models (MLLMs). In the first two baselines, class names are either predefined from extensive vocabularies or extracted using image taggers. The third baseline employs 2D visual tokens, obtained from a pretrained 2D CLIP encoder, which are then processed through a Visual Token Aggregation mechanism before being fed into an LLM to predict class names.
Finally, our proposed method takes this further by consistently aggregating 2D visual tokens from multiview images, transforming them into dense 3D point cloud tokens that enable real-time querying. Notably, our pointwise visual token lifting approach, which feeds these tokens into an LLM, achieves the best performance across multiple benchmarks, matching state-of-the-art OV-3DIS methods that rely on ground-truth class labels.
Furthermore, both our proposed method and designed baselines are entirely training-free, utilizing only pretrained 2D vision encoders (e.g., CLIP \cite{radford2021learning}) and pretrained LLMs (e.g., Vicuna \cite{chiang2023vicuna}). The training-free approach effectively mitigates the prevalent issue of insufficient training data that hampers many existing 3D-LLM techniques.

To evaluate the performance of these methods, we conduct experiments on open-ended versions of two prominent 3D instance segmentation (3DIS) datasets: ScanNet200 \cite{rozenberszki2022language} and ScanNet++ \cite{yeshwanthliu2023scannetpp}. The results underscore the efficacy of our chosen approach over alternative baselines, demonstrating performance levels comparable to OV-3DIS methods that rely on ground-truth class names. Specifically, in ScanNet200, our approach attained an AP of 16.0, contrasting with the 22.2 AP achieved by Open3DIS, currently recognized as the state-of-the-art in OV-3DIS. However, our approach is superior in ScanNet++, where it outperforms Open3DIS by a significant margin (18.4 vs. 13.1 in AP).

In summary, the contributions of our work are:
\begin{enumerate}
    \item We propose Open-Ended 3D Point Cloud Instance Segmentation (OE-3DIS), a task that segments 3D point clouds by instances and generates class names without predefined labels.
    \item We establish solid baselines for OE-3DIS, including leveraging OV-3DIS methods and Multimodal Large Language Models (MLLMs).
    \item We present a training-free OE-3DIS method that lifts 2D visual tokens to 3D and utilizes pretrained Multimodal LLMs to output the final object classes. 
\end{enumerate}

\section{Related Work}
\label{sec:related_work}

\myheading{3D instance segmentation (3DIS)} methods such as Mask3D \cite{Schult23mask3d}, ISBNet \cite{ngo2023isbnet}, PointGroup \cite{jiang2020pointgroup}, and others
\cite{lu2023query, robert2023spt, robert2024scalable, Al_Khatib_2023_spatial} cluster a point cloud scene into 3D instance masks of classes predefined in the training set. 
These methods utilize a 3D Convolutional backbone \cite{choy2019fully, spconv2022, choy2020high} to extract semantic information from the 3D scene. Subsequently, they employ either Dynamic Convolution-based \cite{he2021dyco3d} or Grouping-based \cite{vu2022softgroup} modules to generate 3D instance masks.
Recently, some approaches have adopted techniques to back-project 2D information aggregated from multiple views onto the 3D point cloud to create an ensemble of 3D point cloud features \cite{puy24scalr, Huang2023Segment3D, Peng2023OpenScene, guo2023samgraph}. These 2D-derived features contain rich semantic information, while those derived from 3D capture the geometrical structure of 3D objects. Combined, they supervise a 3D instance decoder to refine segmentation masks. 
However, these methods are closed-vocab or cannot segment new classes in testing, limiting the capability to understand new 3D scenes.

\myheading{Open-vocabulary 3D instance segmentation (OV-3DIS)}
aims at segmenting 3D objects of classes newly provided in testing. To provide 3D proposals for object recognition, OpenMask3D \cite{takmaz2023openmask3d} and Lowis3D \cite{ding2023lowis3d} employ 3DIS networks \cite{ngo2023isbnet,Schult23mask3d} to generate class-agnostic 3D proposals, while SAI3D \cite{yin2023sai3d}, MaskClustering \cite{yan2024maskclustering}, OVIR \cite{lu2023ovir} and Any3DIS \cite{nguyen2025any3dis} utilize 2D segmenter for producing masks for each view and lift these masks to 3D. While 3DIS networks excel in capturing large geometrical structures, they often struggle to detect rare and small-shaped objects. Conversely, 2D segmenters are adept at focusing on small regions but face challenges in maintaining object consistency when lifting to the 3D point cloud. Open3DIS \cite{nguyen2023open3dis} addresses these limitations by combining both 3D and 2D branches, resulting in superior results. This approach effectively captures rare and small objects while preserving the 3D geometrical structures of large objects using superpoint-level masks. 
While OV-3DIS is useful in some scenarios, the constraint of a predefined vocabulary set in inference requires human intervention, hindering very autonomous agents.

\myheading{3D scene understanding with Large Language Models (LLMs).}
Utilizing LLMs for 3D scene understanding focuses on how objects are aligned, their directions, and their locations based on textual questions within 3D environments. This approach emphasizes the spatial aspects of language understanding in three dimensions of data. Previous works \cite{3dllm, huang2023embodied, scanqa, achlioptas2020referit_3d, chen2020scanrefer} have contributed to providing 3D spatial data with language for various applications, including 3D instance and scene captioning \cite{chen2023ll3da, chen2023end, chen2021scan2cap, SpaCap3D, huang2023embodied, 3dvista, 3dllm}, 3D visual answering questions \cite{huang2023embodied, scanqa, ma2022sqa3d, parelli2023clip, 3dvista, 3dllm, delitzas2023multi, chen2023ll3da}, 3D visual grounding \cite{3dvista, 3dllm, huang2023embodied, achlioptas2020referit_3d, chen2020scanrefer, wu2022eda} and supporting embodied AI tasks like planning and reasoning \cite{3dllm, 3dvista, huang2023embodied, chen2023ll3da}.

\myheading{Open-ended 2D Image Understanding} is an emerging task that addresses the need to recognize objects without predefined class names during training or testing. There is scant work on this task, with existing research primarily focusing on image classification \cite{nxtp, conti2023vocabularyfree, zhang2023recognize}, object detection \cite{lin2024generateu}, and instance segmentation \cite{you2023ferret, yu2023towards, Osprey}. Standard 2D Multimodal LLMs (2D MLLMs), such as LLAVA \cite{liu2023llava}, consist of a frozen vision encoder, a projector, and an LLM module. These models typically finetune either (1) the linear projector and the LLM or (2) a complex Q-Former projector. However, applying these methods for 3D scene understanding (3D-LLMs) is challenging due to the lack of sufficient 3D data and text description pairs to effectively train 3D-LLMs. In this paper, we present a novel approach to leveraging pretrained 2D MLLMs for OE-3DIS.

\myheading{OmniScient Model (OSM) \cite{yu2023towards}} is a recently proposed pretrained 2D Multimodal LLM for Open-ended 2D Instance Segmentation, which serves as the foundation for our proposed method.
OSM comprises three main modules: a visual encoder, a MaskQ-Former, and a Large Language Model (LLM).
The visual encoder is a pretrained EVA-CLIP \cite{fang2023eva}, a variant of the CLIP model \cite{radford2021learning}, which extracts high-resolution visual features using a sliding-window scheme and incorporates global positional embeddings to preserve spatial information. The MaskQ-Former, a customized version of Q-Former \cite{instructblip} designed to focus on the mask region rather than the entire image, converts visual features into fixed-length visual tokens. These tokens are then input into the Vicuna LLM \cite{chiang2023vicuna}. The LLM processes these tokens to answer the question, ``What is in the segmentation mask?'' by outputting the object name.

In OSM, only the MaskQ-Former is trained to align visual features with the visual tokens for the LLM, while the visual encoder and LLM remain unchanged. The training datasets are large, including COCO \cite{coco}, LVIS \cite{gupta2019lvis}, ADE20K \cite{ade}, and Cityscapes \cite{cordts2016cityscapes}. This setup demonstrates a strong capability for recognizing objects without predefined class names in 2D images, inspiring us to extend this approach to 3D scene understanding.

\begin{figure*}[t]
  \centering
  \includegraphics[width=0.8\linewidth]{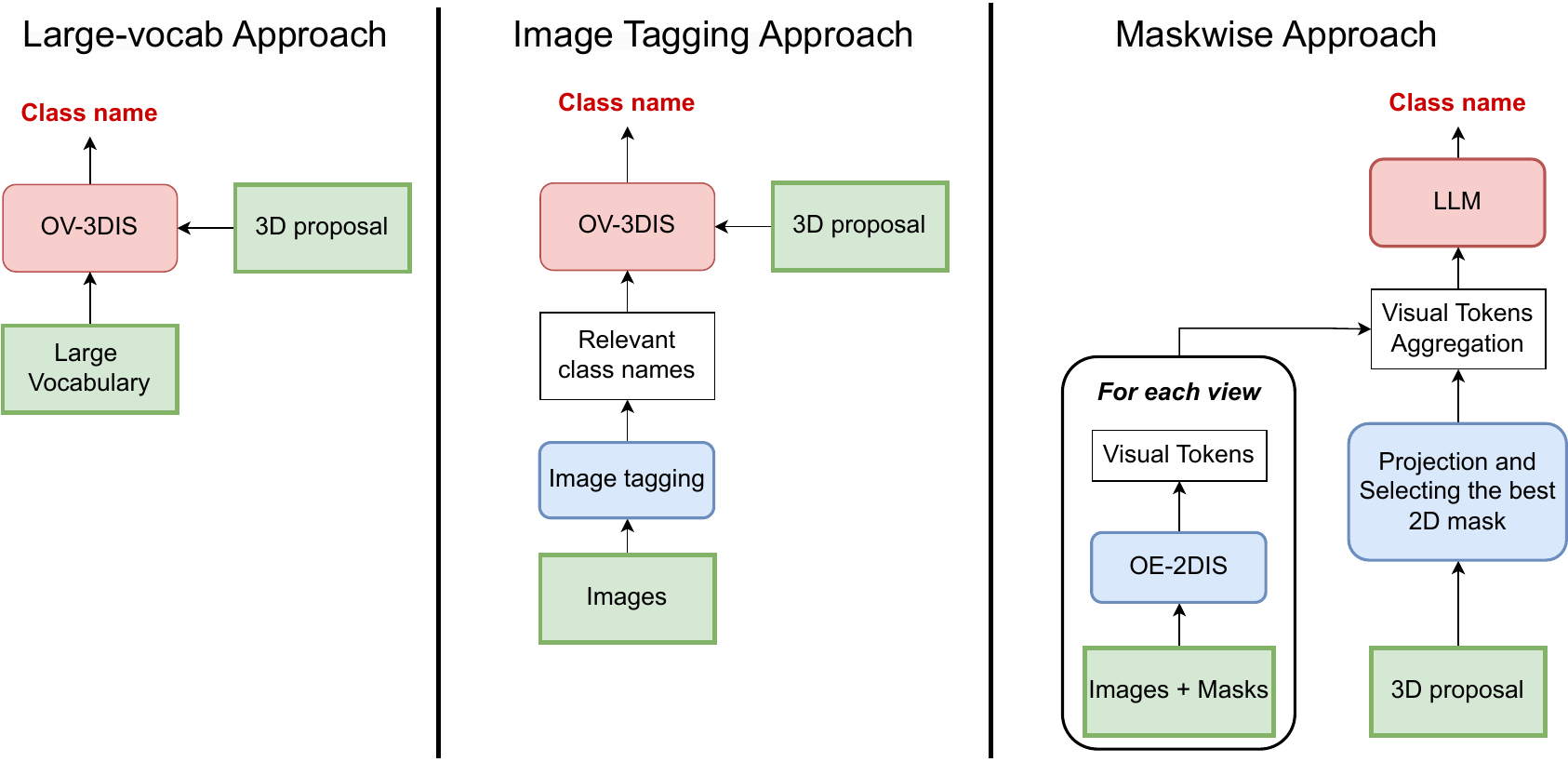}
   \caption{Our baselines: Large-vocab (Left), Image tagging (Middle), and Maskwise (Right).} 
   \vspace{-12pt}
   \label{fig:baseline}
\end{figure*}

\section{Methodology}
\label{sec:method}

\subsection{Problem Statement}
Given a 3D point cloud scene $\mathbf{P} = \{\mathbf{p}_i\}_{i=1}^{N} \in \mathbb{R}^{N \times 6}$ consisting of $N$ points with $xyz$ coordinates and associated $rgb$ colors, along with $T$ RGB-D frames with color images $\{\mathbf{I}_t\}_{t=1}^T$ and depth ones $\{\mathbf{D}_t\}_{t=1}^T$, where $\mathbf{I}_t \in \mathbb{R}^{H\times W \times 3}, \mathbf{D}_t \in \mathbb{R}_+^{H \times W}$, we aim to segment all $K$ object binary masks $\{\mathbf{m}_k\}_{k=1}^{K}$, $\mathbf{m}_k \in \{0, 1\}^N$ and their associated class names $\{l_k\}_{k=1}^{K}$ without giving any predefined class names in inference. Of course, we do need GT class names during evaluation to assess our method's performance. 

The information includes the intrinsic \(\mathbf{\Gamma} \in \mathbb{R}^{3 \times 3}\) and the extrinsic \(\mathbf{[R | v]}_{t} \in \mathbb{R}^{3 \times 4}\). In this context, \(H\) and \(W\) represent the height and width of the image, respectively. The matrix \(\mathbf{R}\) is a 3D rotation matrix, while \(\mathbf{v}\) is a 3D translation vector. This composite matrix, which combines rotation and translation, converts coordinates from the global frame of the point cloud to the camera's frame at view \(t\).

\subsection{Evaluation Metrics}
\label{method:metrics}
To evaluate open-ended object detection or instance segmentation, where predicted class names may be similar but not exactly the same as ground-truth (GT) class names, prior work \cite{Rewatbowornwong2023ZeroGuideSeg} proposed a label reassignment technique. This method uses text encoders (e.g., CLIP \cite{radford2021learning}, BERT \cite{devlin2018bert}, Sentence Transformer \cite{sentence-transformer} to encode both the predicted and GT class names for each scene. It then matches each predicted class name to its closest GT class name based on cosine similarity. After this matching, the standard \textbf{AP score} is used to evaluate performance.

\subsection{Proposed Baselines}
\label{sec:baselines}
Since OE-3DIS is very new and challenging, we focus our efforts on investigating prominent baselines. 
These baselines are illustrated in \cref{fig:baseline}. They require a list of class-agnostic 3D mask proposals pre-extracted from Open3DIS \cite{nguyen2023open3dis} with the DETIC 2D segmenter.

\myheading{Large-vocab approach} (\cref{fig:baseline} - Left): We start with a simple OE-3DIS baseline by using a large vocabulary of 21K common classes introduced by DETIC \cite{detic} as predefined class names for OV-3DIS methods like Open3DIS \cite{nguyen2023open3dis} and OpenMask3D \cite{takmaz2023openmask3d}. However, this approach fails to achieve robust class prediction. This is because the fixed large vocabulary set contains multiple synonyms, resulting in uninformative class predictions after the Softmax operation.

\myheading{Image Tagging approach} (\cref{fig:baseline} - Middle):
To reduce the number of classes, we leverage image-tagging techniques such as RAM++ \cite{zhang2023recognize} to obtain only relevant class names per scene. Specifically, for each input view, a set of image tags is generated and then combined across all processed input views. The resulting unified tag set serves as the vocabulary for OV-3DIS. However, these methods often produce inconsistent class names across views, leading to redundant and similar class names.

\myheading{Maskwise approach} (\cref{fig:baseline} - Right): To tackle the inconsistency in class names across views, we first apply an OE-2DIS method, such as OSM \cite{yu2023towards}, to each view to obtain a list of 2D masks and their predicted fixed-length visual tokens. For each 3D mask proposal, we project it onto a view and associate it with the best-matched 2D mask based on IoU to obtain its 2D fixed-length visual tokens. The 3D visual tokens for the 3D mask are then aggregated by averaging the 2D visual tokens across views, which are subsequently input into a pretrained LLM to obtain the final class names. However, this approach relies on matching 3D proposals with 2D masks, which is often misaligned due to segmentation and depth map imperfections. A 3D mask can project onto multiple 2D masks, and selecting only the best match may discard valuable information.

\begin{figure*}[t]
  \centering
  \includegraphics[width=.9\linewidth]{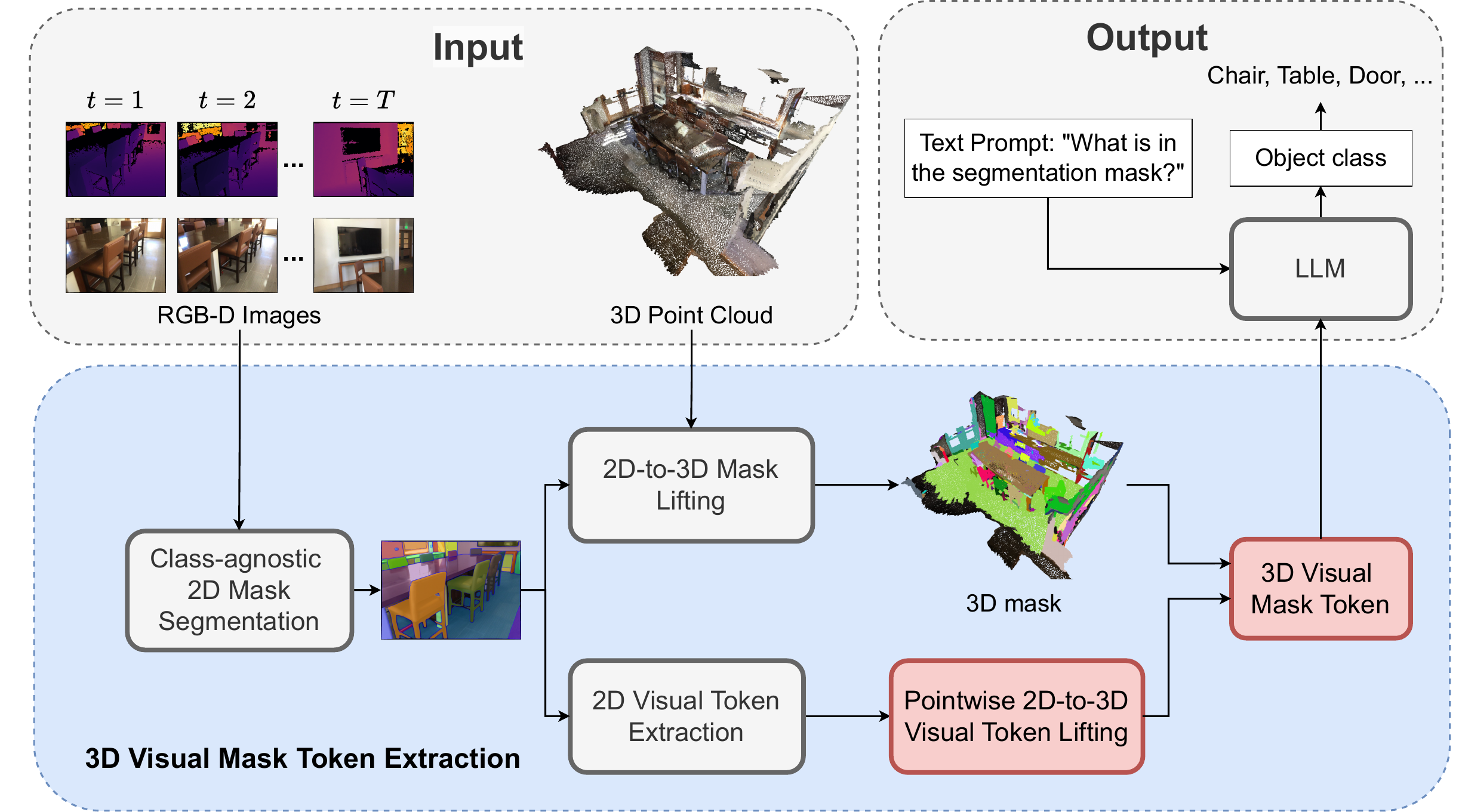}
   \caption{\textbf{Overview of our approach}. First, we generate class-agnostic 2D instance segmentation masks for all views using segmenters like DETIC \cite{detic} and SAM \cite{sam}, and lift these 2D masks into 3D masks using Open3DIS \cite{nguyen2023open3dis}. Simultaneously, the 2D masks and their corresponding RGB images are used to extract 2D visual tokens from an MLLM like OSM \cite{yu2023towards}, which are then lifted into pointwise 3D visual tokens. Finally, for each 3D proposal mask, we aggregate the pointwise 3D visual tokens to form the final tokens for input to the LLM to predict final class names.} 
   \vspace{-12pt}
   \label{fig:architecture}
\end{figure*}

\subsection{Our Approach}
\label{sec:method_pointwise}

To address the above limitations, we propose a method for producing pointwise 3D visual tokens, as illustrated in \cref{fig:architecture}. 
First, we generate class-agnostic 2D instance segmentation masks for all views using class-agnostic 2D segmenters such as DETIC \cite{detic} and SAM \cite{sam}. Next, we lift 2D masks into 3D masks using Open3DIS \cite{nguyen2023open3dis}. Simultaneously, these 2D masks, along with their corresponding RGB images, are used to extract 2D visual tokens from an MLLM like OSM \cite{yu2023towards}. We then lift the resulting 2D visual tokens $\mathbf{F}^{\text{2D}}$ into 3D visual tokens to obtain pointwise 3D visual tokens $\mathbf{F}^{\text{3D}}$. Finally, for each 3D proposal mask, we query the 3D visual tokens associated with this proposal by aggregating the pointwise lifted 3D visual tokens, forming the final tokens $\mathbf{f}^{\text{3D}}$ for input to the LLM.
This approach takes into account the depth and geometric structure of 3D objects via lifting, resulting in more robust visual tokens and a unified, densely-featured point cloud that can be queried instantly at test time. Subsequently, we will focus on our pointwise visual tokens lifting and aggregation. 

Concretely, first, the correspondence of a 3D point $\mathbf{p}_{i} (x, y, z) \in \mathbf{P}$ with its 2D projection $(u, v)$ in view $t$ is:
\begin{equation}
\label{equa:pidepth}
    d_{i,t} \cdot \begin{bmatrix}
u_{i}\\ 
v_{i} \\
1
\end{bmatrix}_t = \mathbf{\Gamma} \cdot \mathbf{[R | c]}_{t} \cdot \begin{bmatrix}
x_{i}\\ 
y_{i}\\ 
z_{i}\\ 
1
\end{bmatrix}
\end{equation}
where $d_{i, t}$ is the projected depth of point $i$ to frame $t$. 

Next, for each view $t$, we extract the 2D visual tokens $\mathbf{F}^{k, \text{2D}}_t \in \mathbb{R}^{E \times C}$, where $E, C$ are the number of visual tokens and feature dimensions, for each 2D mask $k$ from the MaskQ-Former module of OSM \cite{yu2023towards}. Furthermore, we denote $\lambda^k_{t,i} = \{0,1\}$ as the visibility value indicating a point $i$ is visible in mask $k$ of view $t$, $\lambda^k_{t} = \{0,1\}^N$. We set the visibility value $\lambda_{t,i}^k$ of any points whose pixel projections fall outside the $k$-th 2D mask's boundaries or the disparity between projected depth $d$ and the collected depth $\mathbf{D}$ exceeds a defined depth threshold $\tau_{\text{depth}}$, or $|d_{i,t} - \mathbf{D}_{t}[\lfloor u_{i} \rfloor, \lfloor v_{i} \rfloor]| > \tau_{depth}$, to $0$.

Then, we accumulate 3D visual tokens $\mathbf{F}^{\text{3D}} \in \mathbb{R}^{N\times E \times C}$, from every 2D mask $k$ and compute for the frequency $\mathbf{r}^{\text{3D}} \in \mathbb{N}^{N}$ of every view as follows:

\begin{equation}
    \mathbf{F}^{\text{3D}} = \sum_{t, k}\lambda^k_t * \mathbf{F}^{k,\text{2D}}_t, 
    \quad \quad \mathbf{r}^{\text{3D}} = \sum_{t, k} \lambda^k_t ,
\end{equation}
where $*$ is the outer product operation. The normalized pointwise 3D visual tokens are then obtained as follows:
\begin{equation}
    \bar{\mathbf{F}}^{\text{3D}}_i = \begin{cases}
    \mathbf{F}^{\text{3D}}_i / \mathbf{r}^{\text{3D}}_i & \text{if } \mathbf{r}^{\text{3D}}_i > 0 \\
    \mathbf{0} & \text{otherwise}
  \end{cases}. 
\end{equation}

Finally, for each 3D mask $\mathbf{m}_k$, we weighted average the visual tokens of all its points by their frequency to obtain the 3D visual tokens $\mathbf{f}^{\text{3D}}_k$ for that mask, which are then used as input to the LLM to predict the final class, as follows.
\begin{equation}
    \mathbf{f}^{\text{3D}}_k = \frac{\sum_{i \in \mathbf{m}_k} \bar{\mathbf{F}}^{\text{3D}}_i \cdot \mathbf{r}_i^{\text{3D}}}{\sum_{i \in \mathbf{m}_k} \mathbf{r}_i^{\text{3D}}}.
    \label{eq:aggregation}
\end{equation}

\section{Experimental Results}
\label{sec:experiment}

\myheading{Datasets:} We conducted experiments to assess the performance of the baselines and our proposed method using two common 3DIS datasets: ScanNet200 \cite{rozenberszki2022language} and ScanNet++ \cite{yeshwanthliu2023scannetpp}. 
The \textit{ScanNet200} dataset builds on the original ScanNet \cite{dai2017scannet} by expanding its semantic categories from 20 to 200. Its instance segmentation benchmark includes 1,201 training scenes and 312 validation scenes with 198 object categories, significantly enriching the vocabulary and enhancing its capability for detailed 3D semantic and instance segmentation.
The \textit{ScanNet++} dataset was recently introduced, featuring up to 1,659 semantic categories, with 360 training scenes and 50 validation scenes. Given the large number of classes, we follow the standard 3DIS evaluation protocol on ScanNet++ and evaluate only the most common 100 object categories. This dataset offers a much denser 3D point cloud scene representation, making it the most challenging dataset for 3D understanding.

\myheading{Evaluation metrics:} We assess OE-3DIS using the AP score of reassignment of the label (detailed in \cref{method:metrics}). For ScanNet200, we also report AP$_{\text{h}}$ (head), AP$_{\text{c}}$ (common), and AP$_{\text{t}}$ (tail). For ScanNet++, we include the recall rate (RC) and the average recall rate (AR). We note that the OV-3DIS methods adopt a specific AP score calculation protocol by assigning a confidence score of 1.0 to each 3D proposal. Similarly, we follow the same evaluation protocol as the Fully-sup 3DIS by ranking 3D proposals according to their confidence scores. In the context of OE-3DIS, our approach utilizes the confidence scores generated by the LLM, while potential baselines employ the CLIP score.

\myheading{Implementation details:}
\label{sec:implementation}
Following Open3DIS \cite{nguyen2023open3dis}, we generate class-agnostic 3D proposals from ISBNet \cite{ngo2023isbnet} pretrained on ScanNet200 or by lifting 2D masks to 3D using Detic \cite{detic}.
For RAM++ \cite{zhang2023recognize}, we employ the Swin-L model, trained on a 14-million image dataset with an image size of 384px and a tagging threshold of 0.68.
For LLM, we leverage Vicuna-7B \cite{vicuna2023}, fine-tuned for open-ended 2D Instance Segmentation \cite{yu2023towards}.

\begin{table*}[t]
\small
\setlength{\tabcolsep}{4.7pt}
\centering

\begin{tabular}{llcccccccccccc}
\toprule
\multirow{2}{*}{\textbf{Setting}} & \multirow{2}{*}{\textbf{Method}} & \multicolumn{6}{c}{\textbf{ScanNet200}} & \multicolumn{6}{c}{\textbf{ScanNet++}} \\
  \cmidrule(lr){3-8} \cmidrule(lr){9-14} &
   & \textbf{AP} & \textbf{AP$_{50}$} & \textbf{AP$_{25}$}  & \textbf{AP}$_{\text{h}}$ & \textbf{AP}$_{\text{c}}$ & \textbf{AP}$_{\text{t}}$ & \textbf{AP} & \textbf{AP$_{50}$} & \textbf{AP$_{25}$}  & \textbf{AR} & \textbf{RC$_{50}$} & \textbf{RC$_{25}$}  \\ 
\midrule
\rowcolor{gray!30} & Mask3D \cite{Schult23mask3d}  & 26.9 & 36.2 & 41.4 & 39.8 & 21.7 & 17.9 & 8.9 & 14.6 & 38.9 & - & - & - \\
\rowcolor{gray!30} \multirow{-2}{*}{Fully-sup 3DIS} & ISBNet  \cite{ngo2023isbnet}  & 24.5 & 32.7 & 37.6 & 38.6 & 20.5 & 12.5  & 16.7 & 29.7 & 21.0 & - & - & -  \\
\midrule
\rowcolor{gray!30}  & OpenMask3D \cite{takmaz2023openmask3d}  & 15.4 & 19.9 & 23.1 & 17.1 & 14.1 & 14.9 & 2.0 & 2.7 & 3.4 &  4.6 & 8.3 & 12.4  \\
\rowcolor{gray!30} \multirow{-2}{*}{OV-3DIS} & Open3DIS \cite{nguyen2023open3dis} & 23.7 & 29.4 & 32.8 & 27.8 & 21.2 & 21.8 & 13.1 & 20.8 & 24.6 & 22.1 & 33.9 & 39.1 \\
\midrule
Large-Vocab & 21K DETIC classes (Ours)   & 8.5 & 11.7 & 13.1 & 9.9 & 7.2 & 8.3  & 7.3 & 11.9 & 15.2 & 13.3 & 20.3 & 23.6 \\
Image-Tagging & RAM++ \cite{zhang2023recognize} (Ours)  & 10.7 & 14.3 & 16.0 & 11.6 & 11.0 & 9.3 & 9.1 & 15.5 & 19.1 & 16.0 & 24.8 & 28.7 \\

Maskwise& OSM \cite{yu2023towards} (Ours)  & 14.4 & 19.8 & 23.9 & 18.9 & 13.5 & 10.2 &  16.3 & 24.8 & 29.0 & 22.2 & 32.0 & 36.0  \\
\midrule
\multirow{1}{*}{Pointwise} & Ours & \textbf{16.0} & \textbf{22.0} & \textbf{24.7} & \textbf{20.0} & \textbf{14.3} & \textbf{13.2}  & \textbf{18.4} & \textbf{29.4} & \textbf{33.6} & \textbf{23.3} & \textbf{35.2} & \textbf{39.3}\\
\bottomrule

\end{tabular}%
\caption{Comparative results on the ScanNet200 and ScanNet++ datasets. \colorbox{gray!30}{Shaded text} indicates a reference method, not a direct comparison. `-' indicates results are not provided. The best results are in \textbf{bold}.}
\label{tab:scannet200_quanti}
\vspace{-10pt}
\end{table*}

\subsection{Comparison with Baselines}
We compare our approach to the proposed baselines using the OE-3DIS setting on the ScanNet200 and ScanNet++ datasets in \cref{tab:scannet200_quanti}. For reference, we also present results from OV-3DIS methods, including OpenMask3D \cite{takmaz2023openmask3d} and Open3DIS \cite{nguyen2023open3dis}, as well as fully-supervised methods like ISBNet \cite{ngo2023isbnet} and Mask3D \cite{Schult23mask3d} for ScanNet200; and PointGroup \cite{jiang2020pointgroup} and SoftGroup \cite{vu2022softgroup}
for ScanNet++.

\myheading{For Scannet200:} We obtain class-agnostic 3D proposals from two sources: 3D masks from a 3DIS network such as ISBNet \cite{ngo2023isbnet}, and 2D Lift 3D masks from Open3DIS \cite{nguyen2023open3dis}. Our proposed approach outperforms other baselines in the AP score. Furthermore, the performance progression of the baselines, in the specified order, clearly justifies the motivation behind each baseline compared to its predecessor, as discussed in \cref{sec:baselines}. Interestingly, our approach also surpasses OpenMask3D \cite{takmaz2023openmask3d} (16.0 vs. 15.4 in AP), even though OpenMask3D utilizes provided class names. This indicates that, in some cases, we can achieve OV-3DIS without relying on provided class names.

\myheading{For ScanNet++:} Due to the extensive scale and vast array of classes in ScanNet++, the performance of 3D mask results from ISBNet is inadequate. Consequently, we solely rely on the utilization of 2D Lift 3D masks from Open3DIS \cite{nguyen2023open3dis}. We notice a consistent trend akin to the results observed in ScanNet200. Particularly noteworthy is the significant outperformance of our approach compared to OV-3DIS methods or even fully supervised 3DIS, as evidenced by higher AP scores. This underscores the promising application of OE-3DIS in navigating complex 3D scene.

\begin{figure*}[t]
  \centering
  \includegraphics[width=0.85\linewidth]{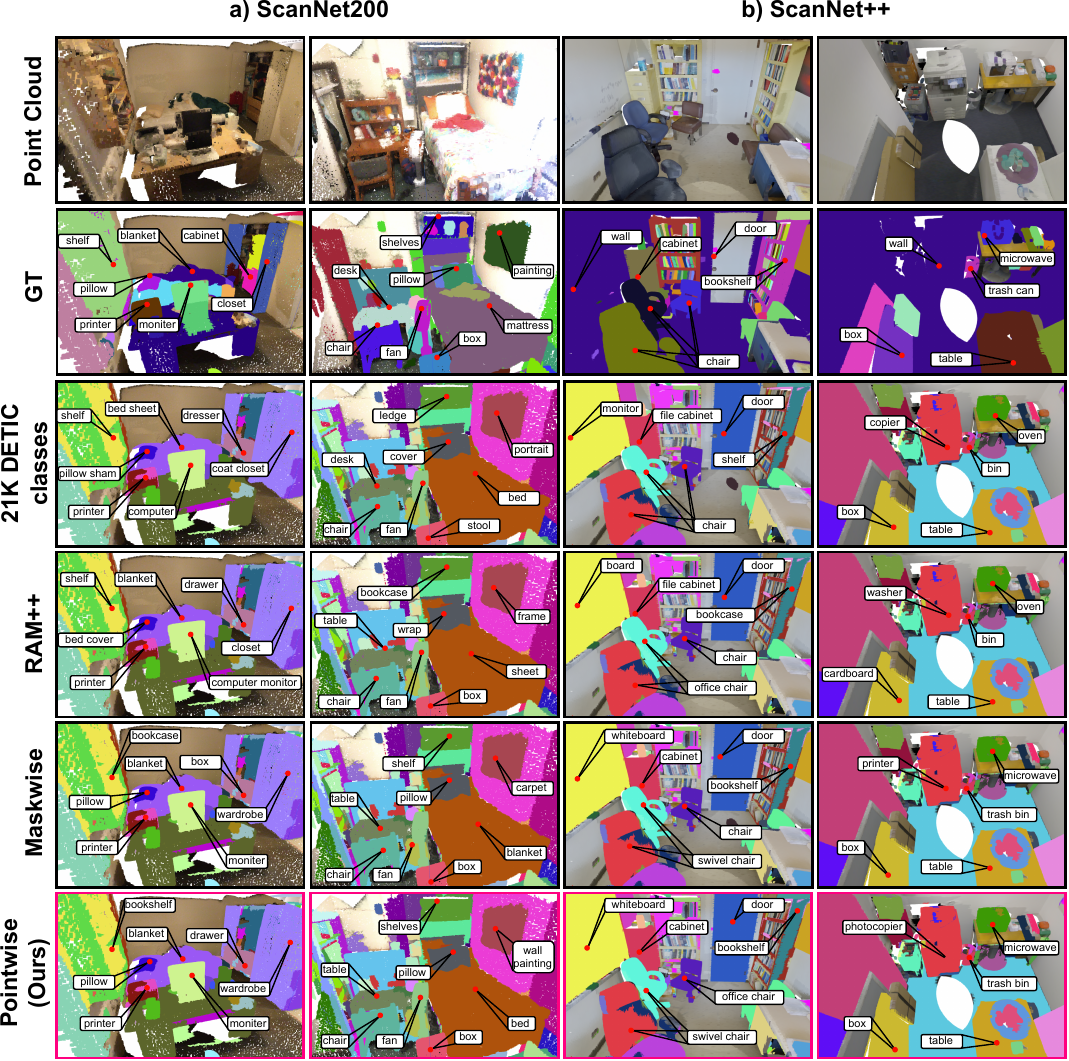}
   \caption{Qualitative examples are provided for ScanNet200 \cite{dai2017scannet} (first two columns) and ScanNet++ \cite{yeshwanthliu2023scannetpp} (last two columns). Both our baselines and approach yield notably good results, particularly in terms of accurately identifying class names even though they do not match exactly the GT classes.} 
   \label{fig:qualitative_scannet200}
\end{figure*}

\myheading{Qualitative comparison:} 
In \cref{fig:qualitative_scannet200}, both our proposed approach and baseline methods effectively assign class names to 3D proposals under OE-3DIS settings. The first column of qualitative results on ScanNet200 demonstrates that our architectures correctly predict class labels for all 3D proposals. The second column highlights notable differences, with our method accurately identifying the `painting' object as a `wall painting', whereas the baselines produce less precise labels. In the final column, our detailed 3D mask proposals surpass the granularity of the ground-truth annotations, enabling accurate recognition of classes absent from ScanNet++'s vocabulary, such as `photocopier'. This illustrates how OE-3DIS allows the model to comprehensively understand a 3D scene without restricting to fixed categories.

\begin{table}[t!]
\small
\setlength{\tabcolsep}{3pt}
\centering

{ %
\begin{tabular}{lcccccc}
\toprule
\textbf{Technique}  & \textbf{AP} & \textbf{AP$_{50}$} & \textbf{AP$_{25}$}  & \textbf{AP}$_{\text{h}}$ & \textbf{AP}$_{\text{c}}$ & \textbf{AP}$_{\text{t}}$ \\ 
\midrule
L2 Norm (Open3DIS) & 5.7 &	8.2 & 9.8 & 6.8 &  4.9  & 5.2 \\
Memory Fusion (OVIR-3D) & 8.4 & 11.6 & 14.4 & 7.2 & 7.0 & 11.6 \\
Max  & 6.3 & 8.4 & 10.2 & 8.5 & 4.7 & 5.6   \\
Random & 13.2 & 18.9 & 21.6 & 17.7 & 12.3 & 10.0  \\
Mean &  14.5 & 20.1 & 22.6 & 18.2 &13.1 & 11.2  \\
\textbf{Weighted Average (Ours)} &   \textbf{16.0} & \textbf{22.0} & \textbf{24.7} & \textbf{20.0} & \textbf{14.3} & \textbf{13.2}  \\

\bottomrule
\end{tabular}%
}
\caption{Ablation on point aggregation techniques}
\label{tab:ablate_aggregation}
\vspace{-10pt}
\end{table}

\begin{table}[t!]
\small
\setlength{\tabcolsep}{3pt}
\centering

{ %
\begin{tabular}{cccccccc}
\toprule
\textbf{3D Proposals} & \textbf{2D Masks} & \textbf{AP} & \textbf{AP$_{50}$}  & \textbf{AP}$_{\text{h}}$ & \textbf{AP}$_{\text{c}}$ & \textbf{AP}$_{\text{t}}$  \\ 
\midrule

3D masks &  DETIC & 11.7 & 16.0  & 16.3 & 10.3 & 8.1  \\
 2D Lift 3D masks & DETIC & 11.7 & 18.6  & 11.3 & 12.1 & 11.8 \\
 3D + 2D Lift 3D masks & DETIC  & \textbf{16.0} & \textbf{22.0}  & \textbf{20.0} & 14.3 & \textbf{13.2} \\
 3D + 2D Lift 3D masks & SAM & 15.4 & 20.5 & 19.2 & \textbf{14.8} & 11.7  \\

\bottomrule
\end{tabular}%
}
\caption{Study on different types 3D proposals.}
\label{tab:ablate_3dproposal}
\end{table}
\begin{table}[t]
\small
\setlength{\tabcolsep}{2pt}
\centering

{ %
\begin{tabular}{lcccccc}
\toprule
\textbf{Text Encoder}  & \textbf{AP} & \textbf{AP$_{50}$} & \textbf{AP$_{25}$}  & \textbf{AP}$_{\text{h}}$ & \textbf{AP}$_{\text{c}}$ & \textbf{AP}$_{\text{t}}$  \\ 
\midrule

BERT \cite{devlin2018bert}  & 13.5 & 19.0 & 21.2 & 17.6 & 11.6 & 11.0  \\
CLIP \cite{radford2021learning} & 15.9 & \textbf{22.0} & 24.5 & \textbf{20.3} & 13.7 & \textbf{13.3}  \\
Sentence Transformer \cite{sentence-transformer} &  \textbf{16.0} & \textbf{22.0} & \textbf{24.7} & 20.0 & \textbf{14.3} & 13.2 \\

\bottomrule
\end{tabular}%
}

\caption{Study on different text encoders in evaluation metrics}
\label{tab:ablate_textencoder}
\end{table}

\begin{table}[t]
\small
\setlength{\tabcolsep}{1pt}
\centering
{ %
\begin{tabular}{lccc}
\toprule
\textbf{Modules}  & \textbf{\# Params(M)} & \textbf{FLOPs (G)} & \textbf{Time (s)}  \\ 
\midrule
Pw. CLIP extract (CLIP)	& 427.94 & 191.11 & 97 \\
QFormer	& 185.66 & 2.74 & 250 \\
LLM (Vicuna 7B) & 6,738.42 & 850.01 & 50 \\
RAM++ & 329.49 & 104.43 & 40 \\
Generate 3D proposals &  N/A &	357.99 & 152 \\
Pw. Visual Token Lift (PVTL) 	& N/A & 4.915 & 314 \\

\bottomrule
\end{tabular}%
}
\caption{Expected latency, FLOPs, and runtime of each module for a 3D scene. Pw. denotes `Pointwise'. Generating 3D proposals from 2D and Pw. visual token lifting are non-parametric modules}
\vspace{-10pt}
\label{tab:params}
\end{table}

\begin{table*}[t]
\small
\setlength{\tabcolsep}{5pt}
\centering
{ %
\begin{tabular}{l cc cccccccc}
\toprule
\textbf{Method} & \textbf{AP} & \textbf{Total Time (s)}	& \textbf{RAM++} & \textbf{Generate 3D proposals} & \textbf{CLIP} & \textbf{QFormer} & \textbf{PVTL} & \textbf{LLM}  \\ 
\midrule
Open3DIS &	N/A &	249 &&\checkmark&\checkmark&&&&		\\
Large Vocab &	8.5	&	249 &&\checkmark&\checkmark&&&& \\
Image Tagging & 10.7 & 289 &\checkmark&\checkmark&\checkmark&&&&	\\
Maskwise & 14.4 & 452 &&\checkmark&&\checkmark&&\checkmark \\
Pointwise & 16.0 & 766 &&\checkmark&&\checkmark&\checkmark&\checkmark \\

\bottomrule
\end{tabular}%
}
\caption{Expected runtime for our proposed baselines and methods with the Open3DIS for a sample 3D scene}
\label{tab:runtime}
\vspace{3pt}
\end{table*}

\begin{table*}[t!]
    \small
    \centering
    \begin{tabular}{p{11.5cm}cc}
    \toprule
    \textbf{Text Prompt} & \bf AP   \\ 
    \midrule
          ``What is in the segmentation mask? Assistant:'' &  \bf 16.0   \\
         ``Can you describe what is in the segmentation mask region? Assistant:''& 15.3   \\
          ``What can you see in the segmentation mask region? Assistant:'' & 15.9   \\
         ``Could you use a few words to describe what is in the segmentation mask? Assistant:'' & 15.2   \\
         ``What is this segmentation mask? Assistant:'' & 15.8 \\

    \bottomrule
    \end{tabular}
    \caption{Study on different text prompts used to query the LLM.}
    \label{tab:ablate_prompting}    
\end{table*}

\subsection{Ablation Study}

To study many design choices of our pointwise approach, we intensively carry out ablation study on the ScanNet200 \cite{dai2017scannet} dataset.

\myheading{Study on different point feature aggregation techniques (\cref{eq:aggregation}}) is shown in \cref{tab:ablate_aggregation}. We evaluate four techniques for combining point features from a given 3D mask: L2-norm, max, random (randomly selecting one point), mean, and weighted average (our proposed operation). The weighted average technique achieved the highest performance with an AP score of 16.0, outperforming the other methods. Consequently, we utilized the weighted average technique in all our experiments.

\myheading{Study on different types 3D proposals.} As reported in \cref{tab:ablate_3dproposal}, combining 3D proposals from both 3D masks and 2D Lift 3D masks yielded the most favorable results. This outcome validates our choice of 3D proposals for the ScanNet200 dataset. Additionally, the 2D Lift 3D proposals obtained from DETIC demonstrated slightly superior performance compared to those from SAM. This observation diverges from the methodology adopted in Open3DIS \cite{nguyen2023open3dis}, where provided class names are utilized to employ Grounded-SAM \cite{ren2024grounded} instead of SAM \cite{sam}, as in our OE-3DIS scenario.

\myheading{Study on different text encoders for evaluation metrics} is described in \cref{tab:ablate_textencoder}.We evaluate three text decoders: BERT \cite{devlin2018bert}, CLIP \cite{radford2021learning}, and Sentence Transformer \cite{sentence-transformer}. Among these, Sentence Transformer embeddings achieve the highest AP scores, outperforming CLIP and BERT, which yield comparatively lower scores. We also observe that CLIP embeddings exhibit greater similarity among themselves compared to those from BERT. Moreover, Sentence Transformer embeddings are better suited for capturing complex sentence-level descriptions rather than focusing solely on individual class names. Therefore, we recommend Sentence Transformer as our preferred text encoder.

\myheading{Latency analysis of our baselines and proposed method} is presented in \cref{tab:params} (excluding the runtime for generating 2D proposals). Additionally, \cref{tab:runtime} compares the total runtime across our baselines, proposed methods, and the state-of-the-art 3DIS approach, Open3DIS. As shown, our method prioritizes accuracy at the cost of increased runtime, meaning higher accuracy comes with slower execution.

\myheading{Study on different text prompts used to query the LLM}: We experiment with various input text prompts to query the LLM for class names. \cref{tab:ablate_prompting} demonstrates that altering the prompt subtly affects the model's accuracy. We select the prompt ``What is in the segmentation mask? Assistant:'' from the variants for our approach.

\section{Discussion and Conclusion}
\label{sec:conclusion}
\myheading{Limitations:}
The motivation behind our proposed Open-ended 3D Instance Segmentation (OE-3DIS) arises from limitations in current Open-Vocabulary 3D Point Cloud Instance Segmentation (OV-3DIS) methods, which still depend on a predefined set of class names during testing. This constraint is impractical in scenarios lacking prior knowledge of class names, such as a robot navigating unfamiliar environments. Although our method advances open-ended 3D scene understanding, it still faces notable limitations. Firstly, our approach heavily depends on 2D visual tokens extracted from a pretrained OSM, which itself is trained on instance segmentation datasets containing only a limited number of classes. Consequently, this restricts the model’s capability to recognize an extensive range of classes in truly open-world contexts. Secondly, the performance of OE-3DIS relies significantly on class-agnostic 3DIS methods, whose effectiveness in turn depends on the accuracy of 3D representations and the quality of 2D-to-3D mapping, including factors like camera calibration and depth image quality.

\myheading{Conclusion:}
We have introduced Open-Ended 3D Point Cloud Instance Segmentation (OE-3DIS), which generates 3D masks and object class names without predefined labels during testing. We have explored baselines using OV-3DIS methods and MLLMs, and introduced a pointwise training-free approach leveraging OSM. Experiments on ScanNet200 and ScanNet++ show our approach’s superior performance, notably outperforming Open3DIS (SOTA on OV-3DIS) on ScanNet++ without ground-truth class names.

{
    \small
    \bibliographystyle{ieeenat_fullname}
    \bibliography{main}

\begin{thebibliography}{64}
\providecommand{\natexlab}[1]{#1}
\providecommand{\url}[1]{\texttt{#1}}
\expandafter\ifx\csname urlstyle\endcsname\relax
  \providecommand{\doi}[1]{doi: #1}\else
  \providecommand{\doi}{doi: \begingroup \urlstyle{rm}\Url}\fi

\bibitem[Achlioptas et~al.(2020)Achlioptas, Abdelreheem, Xia, Elhoseiny, and Guibas]{achlioptas2020referit_3d}
Panos Achlioptas, Ahmed Abdelreheem, Fei Xia, Mohamed Elhoseiny, and Leonidas~J. Guibas.
\newblock {ReferIt3D}: Neural listeners for fine-grained 3d object identification in real-world scenes.
\newblock In \emph{16th European Conference on Computer Vision (ECCV)}, 2020.

\bibitem[Al~Khatib et~al.(2023)Al~Khatib, El~Amine~Boudjoghra, Lahoud, and Khan]{Al_Khatib_2023_spatial}
Salwa Al~Khatib, Mohamed El~Amine~Boudjoghra, Jean Lahoud, and Fahad~Shahbaz Khan.
\newblock 3d instance segmentation via enhanced spatial and semantic supervision.
\newblock In \emph{Proceedings of the IEEE/CVF International Conference on Computer Vision (ICCV)}, pages 541--550, 2023.

\bibitem[Azuma et~al.(2022)Azuma, Miyanishi, Kurita, and Kawanabe]{scanqa}
Daichi Azuma, Taiki Miyanishi, Shuhei Kurita, and Motoaki Kawanabe.
\newblock Scanqa: 3d question answering for spatial scene understanding.
\newblock In \emph{Proceedings of the IEEE/CVF Conference on Computer Vision and Pattern Recognition (CVPR)}, 2022.

\bibitem[Chen et~al.(2020)Chen, Chang, and Nie{\ss}ner]{chen2020scanrefer}
Dave~Zhenyu Chen, Angel~X Chang, and Matthias Nie{\ss}ner.
\newblock Scanrefer: 3d object localization in rgb-d scans using natural language.
\newblock \emph{16th European Conference on Computer Vision (ECCV)}, 2020.

\bibitem[Chen et~al.(2021{\natexlab{a}})Chen, Fang, Zhang, Liu, and Wang]{chen2021hierarchical}
Shaoyu Chen, Jiemin Fang, Qian Zhang, Wenyu Liu, and Xinggang Wang.
\newblock Hierarchical aggregation for 3d instance segmentation.
\newblock In \emph{Proceedings of the IEEE/CVF International Conference on Computer Vision}, pages 15467--15476, 2021{\natexlab{a}}.

\bibitem[Chen et~al.(2023{\natexlab{a}})Chen, Chen, Zhang, Li, Yu, Fei, Zhu, Fan, and Chen]{chen2023ll3da}
Sijin Chen, Xin Chen, Chi Zhang, Mingsheng Li, Gang Yu, Hao Fei, Hongyuan Zhu, Jiayuan Fan, and Tao Chen.
\newblock Ll3da: Visual interactive instruction tuning for omni-3d understanding, reasoning, and planning, 2023{\natexlab{a}}.

\bibitem[Chen et~al.(2023{\natexlab{b}})Chen, Zhu, Chen, Lei, Yu, and Chen]{chen2023end}
Sijin Chen, Hongyuan Zhu, Xin Chen, Yinjie Lei, Gang Yu, and Tao Chen.
\newblock End-to-end 3d dense captioning with vote2cap-detr.
\newblock In \emph{Proceedings of the IEEE/CVF Conference on Computer Vision and Pattern Recognition}, pages 11124--11133, 2023{\natexlab{b}}.

\bibitem[Chen et~al.(2021{\natexlab{b}})Chen, Gholami, Nie{\ss}ner, and Chang]{chen2021scan2cap}
Zhenyu Chen, Ali Gholami, Matthias Nie{\ss}ner, and Angel~X Chang.
\newblock Scan2cap: Context-aware dense captioning in rgb-d scans.
\newblock In \emph{Proceedings of the IEEE/CVF Conference on Computer Vision and Pattern Recognition}, pages 3193--3203, 2021{\natexlab{b}}.

\bibitem[Chiang et~al.(2023{\natexlab{a}})Chiang, Li, Lin, Sheng, Wu, Zhang, Zheng, Zhuang, Zhuang, Gonzalez, Stoica, and Xing]{vicuna2023}
Wei-Lin Chiang, Zhuohan Li, Zi Lin, Ying Sheng, Zhanghao Wu, Hao Zhang, Lianmin Zheng, Siyuan Zhuang, Yonghao Zhuang, Joseph~E. Gonzalez, Ion Stoica, and Eric~P. Xing.
\newblock Vicuna: An open-source chatbot impressing gpt-4 with 90\%* chatgpt quality, 2023{\natexlab{a}}.

\bibitem[Chiang et~al.(2023{\natexlab{b}})Chiang, Li, Lin, Sheng, Wu, Zhang, Zheng, Zhuang, Zhuang, Gonzalez, et~al.]{chiang2023vicuna}
Wei-Lin Chiang, Zhuohan Li, Zi Lin, Ying Sheng, Zhanghao Wu, Hao Zhang, Lianmin Zheng, Siyuan Zhuang, Yonghao Zhuang, Joseph~E Gonzalez, et~al.
\newblock Vicuna: An open-source chatbot impressing gpt-4 with 90\%* chatgpt quality.
\newblock \emph{See https://vicuna. lmsys. org (accessed 14 April 2023)}, 2\penalty0 (3):\penalty0 6, 2023{\natexlab{b}}.

\bibitem[Choy et~al.(2019)Choy, Park, and Koltun]{choy2019fully}
Christopher Choy, Jaesik Park, and Vladlen Koltun.
\newblock Fully convolutional geometric features.
\newblock In \emph{Proceedings of the IEEE International Conference on Computer Vision}, pages 8958--8966, 2019.

\bibitem[Choy et~al.(2020)Choy, Lee, Ranftl, Park, and Koltun]{choy2020high}
Christopher Choy, Junha Lee, Rene Ranftl, Jaesik Park, and Vladlen Koltun.
\newblock High-dimensional convolutional networks for geometric pattern recognition.
\newblock In \emph{Proceedings of the IEEE Conference on Computer Vision and Pattern Recognition}, 2020.

\bibitem[Chuang et~al.(2024)Chuang, Yi, Lizhen, Zehuan, and Jianfei]{lin2024generateu}
Lin Chuang, Jiang Yi, Qu Lizhen, Yuan Zehuan, and Cai Jianfei.
\newblock Generative region-language pretraining for open-ended object detection.
\newblock In \emph{Proceedings of IEEE Conference on Computer Vision and Pattern Recognition (CVPR)}, 2024.

\bibitem[Conti et~al.(2023)Conti, Fini, Mancini, Rota, Wang, and Ricci]{conti2023vocabularyfree}
Alessandro Conti, Enrico Fini, Massimiliano Mancini, Paolo Rota, Yiming Wang, and Elisa Ricci.
\newblock Vocabulary-free image classification, 2023.

\bibitem[Contributors(2022)]{spconv2022}
Spconv Contributors.
\newblock Spconv: Spatially sparse convolution library.
\newblock \url{https://github.com/traveller59/spconv}, 2022.

\bibitem[Cordts et~al.(2016)Cordts, Omran, Ramos, Rehfeld, Enzweiler, Benenson, Franke, Roth, and Schiele]{cordts2016cityscapes}
Marius Cordts, Mohamed Omran, Sebastian Ramos, Timo Rehfeld, Markus Enzweiler, Rodrigo Benenson, Uwe Franke, Stefan Roth, and Bernt Schiele.
\newblock The cityscapes dataset for semantic urban scene understanding.
\newblock In \emph{Proceedings of the IEEE conference on computer vision and pattern recognition}, pages 3213--3223, 2016.

\bibitem[Dai et~al.(2017)Dai, Chang, Savva, Halber, Funkhouser, and Nie{\ss}ner]{dai2017scannet}
Angela Dai, Angel~X. Chang, Manolis Savva, Maciej Halber, Thomas Funkhouser, and Matthias Nie{\ss}ner.
\newblock Scannet: Richly-annotated 3d reconstructions of indoor scenes.
\newblock In \emph{Proc. Computer Vision and Pattern Recognition (CVPR), IEEE}, 2017.

\bibitem[Dai et~al.(2023)Dai, Li, Li, Tiong, Zhao, Wang, Li, Fung, and Hoi]{instructblip}
Wenliang Dai, Junnan Li, Dongxu Li, Anthony Meng~Huat Tiong, Junqi Zhao, Weisheng Wang, Boyang Li, Pascale Fung, and Steven Hoi.
\newblock Instructblip: Towards general-purpose vision-language models with instruction tuning, 2023.

\bibitem[Delitzas et~al.(2023)Delitzas, Parelli, Hars, Vlassis, Anagnostidis, Bachmann, and Hofmann]{delitzas2023multi}
Alexandros Delitzas, Maria Parelli, Nikolas Hars, Georgios Vlassis, Sotirios Anagnostidis, Gregor Bachmann, and Thomas Hofmann.
\newblock Multi-clip: Contrastive vision-language pre-training for question answering tasks in 3d scenes.
\newblock \emph{arXiv preprint arXiv:2306.02329}, 2023.

\bibitem[Devlin et~al.(2018)Devlin, Chang, Lee, and Toutanova]{devlin2018bert}
Jacob Devlin, Ming-Wei Chang, Kenton Lee, and Kristina Toutanova.
\newblock Bert: Pre-training of deep bidirectional transformers for language understanding.
\newblock \emph{arXiv preprint arXiv:1810.04805}, 2018.

\bibitem[Ding et~al.(2023)Ding, Yang, Xue, Zhang, Bai, and Qi]{ding2023lowis3d}
Runyu Ding, Jihan Yang, Chuhui Xue, Wenqing Zhang, Song Bai, and Xiaojuan Qi.
\newblock Lowis3d: Language-driven open-world instance-level 3d scene understanding, 2023.

\bibitem[Fang et~al.(2023)Fang, Wang, Xie, Sun, Wu, Wang, Huang, Wang, and Cao]{fang2023eva}
Yuxin Fang, Wen Wang, Binhui Xie, Quan Sun, Ledell Wu, Xinggang Wang, Tiejun Huang, Xinlong Wang, and Yue Cao.
\newblock Eva: Exploring the limits of masked visual representation learning at scale.
\newblock In \emph{Proceedings of the IEEE/CVF Conference on Computer Vision and Pattern Recognition}, pages 19358--19369, 2023.

\bibitem[Guo et~al.(2023)Guo, Zhu, Peng, Wang, Shen, Hu, and Zhou]{guo2023samgraph}
Haoyu Guo, He Zhu, Sida Peng, Yuang Wang, Yujun Shen, Ruizhen Hu, and Xiaowei Zhou.
\newblock Sam-guided graph cut for 3d instance segmentation.
\newblock \emph{arXiv preprint arXiv:2312.08372}, 2023.

\bibitem[Gupta et~al.(2019)Gupta, Dollar, and Girshick]{gupta2019lvis}
Agrim Gupta, Piotr Dollar, and Ross Girshick.
\newblock Lvis: A dataset for large vocabulary instance segmentation.
\newblock In \emph{Proceedings of the IEEE/CVF conference on computer vision and pattern recognition}, pages 5356--5364, 2019.

\bibitem[He et~al.(2021)He, Shen, and van~den Hengel]{he2021dyco3d}
Tong He, Chunhua Shen, and Anton van~den Hengel.
\newblock Dyco3d: Robust instance segmentation of 3d point clouds through dynamic convolution.
\newblock In \emph{Proceedings of the IEEE/CVF Conference on Computer Vision and Pattern Recognition}, pages 354--363, 2021.

\bibitem[Hong et~al.(2023)Hong, Zhen, Chen, Zheng, Du, Chen, and Gan]{3dllm}
Yining Hong, Haoyu Zhen, Peihao Chen, Shuhong Zheng, Yilun Du, Zhenfang Chen, and Chuang Gan.
\newblock 3d-llm: Injecting the 3d world into large language models.
\newblock \emph{NeurIPS}, 2023.

\bibitem[Huang et~al.(2024)Huang, Yong, Ma, Linghu, Li, Wang, Li, Zhu, Jia, and Huang]{huang2023embodied}
Jiangyong Huang, Silong Yong, Xiaojian Ma, Xiongkun Linghu, Puhao Li, Yan Wang, Qing Li, Song-Chun Zhu, Baoxiong Jia, and Siyuan Huang.
\newblock An embodied generalist agent in 3d world.
\newblock In \emph{Proceedings of the International Conference on Machine Learning (ICML)}, 2024.

\bibitem[Huang et~al.(2023{\natexlab{a}})Huang, Peng, Takmaz, Tombari, Pollefeys, Song, Huang, and Engelmann]{Huang2023Segment3D}
Rui Huang, Songyou Peng, Ayca Takmaz, Federico Tombari, Marc Pollefeys, Shiji Song, Gao Huang, and Francis Engelmann.
\newblock Segment3d: Learning fine-grained class-agnostic 3d segmentation without manual labels.
\newblock \emph{arXiv}, 2023{\natexlab{a}}.

\bibitem[Huang et~al.(2023{\natexlab{b}})Huang, Huang, Zhang, Tian, Feng, Zhang, Xie, Li, and Zhang]{zhang2023recognize}
Xinyu Huang, Yi-Jie Huang, Youcai Zhang, Weiwei Tian, Rui Feng, Yuejie Zhang, Yanchun Xie, Yaqian Li, and Lei Zhang.
\newblock Open-set image tagging with multi-grained text supervision, 2023{\natexlab{b}}.

\bibitem[Jiang et~al.(2020)Jiang, Zhao, Shi, Liu, Fu, and Jia]{jiang2020pointgroup}
Li Jiang, Hengshuang Zhao, Shaoshuai Shi, Shu Liu, Chi-Wing Fu, and Jiaya Jia.
\newblock Pointgroup: Dual-set point grouping for 3d instance segmentation.
\newblock In \emph{Proceedings of the IEEE/CVF Conference on Computer Vision and Pattern Recognition}, pages 4867--4876, 2020.

\bibitem[Kirillov et~al.(2023)Kirillov, Mintun, Ravi, Mao, Rolland, Gustafson, Xiao, Whitehead, Berg, Lo, Doll{\'a}r, and Girshick]{sam}
Alexander Kirillov, Eric Mintun, Nikhila Ravi, Hanzi Mao, Chloe Rolland, Laura Gustafson, Tete Xiao, Spencer Whitehead, Alexander~C. Berg, Wan-Yen Lo, Piotr Doll{\'a}r, and Ross Girshick.
\newblock Segment anything.
\newblock \emph{arXiv:2304.02643}, 2023.

\bibitem[Lin et~al.(2014)Lin, Maire, Belongie, Hays, Perona, Ramanan, Doll{\'a}r, and Zitnick]{coco}
Tsung-Yi Lin, Michael Maire, Serge Belongie, James Hays, Pietro Perona, Deva Ramanan, Piotr Doll{\'a}r, and C~Lawrence Zitnick.
\newblock Microsoft coco: Common objects in context.
\newblock In \emph{Computer Vision--ECCV 2014: 13th European Conference, Zurich, Switzerland, September 6-12, 2014, Proceedings, Part V 13}, pages 740--755. Springer, 2014.

\bibitem[Liu et~al.(2023)Liu, Li, Wu, and Lee]{liu2023llava}
Haotian Liu, Chunyuan Li, Qingyang Wu, and Yong~Jae Lee.
\newblock Visual instruction tuning.
\newblock In \emph{NeurIPS}, 2023.

\bibitem[Lu et~al.(2023{\natexlab{a}})Lu, Deng, Wang, He, and Zhang]{lu2023query}
Jiahao Lu, Jiacheng Deng, Chuxin Wang, Jianfeng He, and Tianzhu Zhang.
\newblock Query refinement transformer for 3d instance segmentation.
\newblock In \emph{Proceedings of the IEEE/CVF International Conference on Computer Vision (ICCV)}, pages 18516--18526, 2023{\natexlab{a}}.

\bibitem[Lu et~al.(2023{\natexlab{b}})Lu, Chang, Jing, Boularias, and Bekris]{lu2023ovir}
Shiyang Lu, Haonan Chang, Eric~Pu Jing, Abdeslam Boularias, and Kostas Bekris.
\newblock Ovir-3d: Open-vocabulary 3d instance retrieval without training on 3d data.
\newblock In \emph{7th Annual Conference on Robot Learning}, 2023{\natexlab{b}}.

\bibitem[Ma et~al.(2023)Ma, Yong, Zheng, Li, Liang, Zhu, and Huang]{ma2022sqa3d}
Xiaojian Ma, Silong Yong, Zilong Zheng, Qing Li, Yitao Liang, Song-Chun Zhu, and Siyuan Huang.
\newblock Sqa3d: Situated question answering in 3d scenes.
\newblock In \emph{International Conference on Learning Representations}, 2023.

\bibitem[Ngo et~al.(2023)Ngo, Hua, and Nguyen]{ngo2023isbnet}
Tuan~Duc Ngo, Binh-Son Hua, and Khoi Nguyen.
\newblock Isbnet: a 3d point cloud instance segmentation network with instance-aware sampling and box-aware dynamic convolution.
\newblock In \emph{Proceedings of the IEEE/CVF Conference on Computer Vision and Pattern Recognition}, pages 13550--13559, 2023.

\bibitem[Nguyen et~al.(2025)Nguyen, Luu, Tran, Pham, and Nguyen]{nguyen2025any3dis}
Phuc Nguyen, Minh Luu, Anh Tran, Cuong Pham, and Khoi Nguyen.
\newblock Any3dis: Class-agnostic 3d instance segmentation by 2d mask tracking.
\newblock In \emph{Proceedings of the Computer Vision and Pattern Recognition Conference}, pages 3636--3645, 2025.

\bibitem[Nguyen et~al.(2024)Nguyen, Ngo, Kalogerakis, Gan, Tran, Pham, and Nguyen]{nguyen2023open3dis}
Phuc D.~A. Nguyen, Tuan~Duc Ngo, Evangelos Kalogerakis, Chuang Gan, Anh Tran, Cuong Pham, and Khoi Nguyen.
\newblock Open3dis: Open-vocabulary 3d instance segmentation with 2d mask guidance.
\newblock In \emph{Proceedings of the IEEE/CVF Conference on Computer Vision and Pattern Recognition (CVPR)}, 2024.

\bibitem[Parelli et~al.(2023)Parelli, Delitzas, Hars, Vlassis, Anagnostidis, Bachmann, and Hofmann]{parelli2023clip}
Maria Parelli, Alexandros Delitzas, Nikolas Hars, Georgios Vlassis, Sotirios Anagnostidis, Gregor Bachmann, and Thomas Hofmann.
\newblock Clip-guided vision-language pre-training for question answering in 3d scenes.
\newblock In \emph{Proceedings of the IEEE/CVF Conference on Computer Vision and Pattern Recognition}, pages 5606--5611, 2023.

\bibitem[Peng et~al.(2023)Peng, Genova, Jiang, Tagliasacchi, Pollefeys, and Funkhouser]{Peng2023OpenScene}
Songyou Peng, Kyle Genova, Chiyu~"Max" Jiang, Andrea Tagliasacchi, Marc Pollefeys, and Thomas Funkhouser.
\newblock Openscene: 3d scene understanding with open vocabularies.
\newblock In \emph{Proceedings of the IEEE/CVF Conference on Computer Vision and Pattern Recognition (CVPR)}, 2023.

\bibitem[Puy et~al.(2024)Puy, Gidaris, Boulch, Sim\'eoni, Sautier, P\'erez, Bursuc, and Marlet]{puy24scalr}
Gilles Puy, Spyros Gidaris, Alexandre Boulch, Oriane Sim\'eoni, Corentin Sautier, Patrick P\'erez, Andrei Bursuc, and Renaud Marlet.
\newblock Three pillars improving vision foundation model distillation for lidar.
\newblock In \emph{CVPR}, 2024.

\bibitem[Radford et~al.(2021)Radford, Kim, Hallacy, Ramesh, Goh, Agarwal, Sastry, Askell, Mishkin, Clark, et~al.]{radford2021learning}
Alec Radford, Jong~Wook Kim, Chris Hallacy, Aditya Ramesh, Gabriel Goh, Sandhini Agarwal, Girish Sastry, Amanda Askell, Pamela Mishkin, Jack Clark, et~al.
\newblock Learning transferable visual models from natural language supervision.
\newblock In \emph{International conference on machine learning}, pages 8748--8763. PMLR, 2021.

\bibitem[Reimers and Gurevych(2020)]{sentence-transformer}
Nils Reimers and Iryna Gurevych.
\newblock Making monolingual sentence embeddings multilingual using knowledge distillation.
\newblock In \emph{Proceedings of the 2020 Conference on Empirical Methods in Natural Language Processing}. Association for Computational Linguistics, 2020.

\bibitem[Ren et~al.(2024)Ren, Liu, Zeng, Lin, Li, Cao, Chen, Huang, Chen, Yan, et~al.]{ren2024grounded}
Tianhe Ren, Shilong Liu, Ailing Zeng, Jing Lin, Kunchang Li, He Cao, Jiayu Chen, Xinyu Huang, Yukang Chen, Feng Yan, et~al.
\newblock Grounded sam: Assembling open-world models for diverse visual tasks.
\newblock \emph{arXiv preprint arXiv:2401.14159}, 2024.

\bibitem[Rewatbowornwong et~al.(2023)Rewatbowornwong, Chatthee, Chuangsuwanich, and Suwajanakorn]{Rewatbowornwong2023ZeroGuideSeg}
Pitchaporn Rewatbowornwong, Nattanat Chatthee, Ekapol Chuangsuwanich, and Supasorn Suwajanakorn.
\newblock Zero-guidance segmentation using zero segment labels.
\newblock In \emph{IEEE International Conference on Computer Vision (ICCV)}, 2023.

\bibitem[Robert et~al.(2023)Robert, Raguet, and Landrieu]{robert2023spt}
Damien Robert, Hugo Raguet, and Loic Landrieu.
\newblock Efficient 3d semantic segmentation with superpoint transformer.
\newblock \emph{Proceedings of the IEEE/CVF International Conference on Computer Vision}, 2023.

\bibitem[Robert et~al.(2024)Robert, Raguet, and Landrieu]{robert2024scalable}
Damien Robert, Hugo Raguet, and Loic Landrieu.
\newblock Scalable 3d panoptic segmentation as superpoint graph clustering.
\newblock \emph{Proceedings of the IEEE International Conference on 3D Vision}, 2024.

\bibitem[Rozenberszki et~al.(2022)Rozenberszki, Litany, and Dai]{rozenberszki2022language}
David Rozenberszki, Or Litany, and Angela Dai.
\newblock Language-grounded indoor 3d semantic segmentation in the wild.
\newblock In \emph{Proceedings of the European Conference on Computer Vision ({ECCV})}, 2022.

\bibitem[Schult et~al.(2023)Schult, Engelmann, Hermans, Litany, Tang, and Leibe]{Schult23mask3d}
Jonas Schult, Francis Engelmann, Alexander Hermans, Or Litany, Siyu Tang, and Bastian Leibe.
\newblock Mask3d for 3d semantic instance segmentation.
\newblock In \emph{{International Conference on Robotics and Automation (ICRA)}}, 2023.

\bibitem[Takmaz et~al.(2023)Takmaz, Fedele, Sumner, Pollefeys, Tombari, and Engelmann]{takmaz2023openmask3d}
Ay{\c{c}}a Takmaz, Elisabetta Fedele, Robert~W. Sumner, Marc Pollefeys, Federico Tombari, and Francis Engelmann.
\newblock {OpenMask3D: Open-Vocabulary 3D Instance Segmentation}.
\newblock In \emph{Advances in Neural Information Processing Systems (NeurIPS)}, 2023.

\bibitem[Vu et~al.(2022)Vu, Kim, Luu, Nguyen, and Yoo]{vu2022softgroup}
Thang Vu, Kookhoi Kim, Tung~M. Luu, Xuan~Thanh Nguyen, and Chang~D. Yoo.
\newblock Softgroup for 3d instance segmentation on 3d point clouds.
\newblock In \emph{CVPR}, 2022.

\bibitem[Wang et~al.(2022)Wang, Zhang, Yu, and Cai]{SpaCap3D}
Heng Wang, Chaoyi Zhang, Jianhui Yu, and Weidong Cai.
\newblock Spatiality-guided transformer for 3{D} dense captioning on point clouds.
\newblock In \emph{Proceedings of the Thirty-First International Joint Conference on Artificial Intelligence, {IJCAI-22}}, 2022.

\bibitem[Wu et~al.(2023)Wu, Cheng, Zhang, Cheng, and Zhang]{wu2022eda}
Yanmin Wu, Xinhua Cheng, Renrui Zhang, Zesen Cheng, and Jian Zhang.
\newblock Eda: Explicit text-decoupling and dense alignment for 3d visual grounding.
\newblock In \emph{Proceedings of the IEEE Conference on Computer Vision and Pattern Recognition (CVPR)}, 2023.

\bibitem[Yan et~al.(2024)Yan, Zhang, Zhu, and Wang]{yan2024maskclustering}
Mi Yan, Jiazhao Zhang, Yan Zhu, and He Wang.
\newblock Maskclustering: View consensus based mask graph clustering for open-vocabulary 3d instance segmentation.
\newblock \emph{arXiv preprint arXiv:2401.07745}, 2024.

\bibitem[Yeshwanth et~al.(2023)Yeshwanth, Liu, Nie{\ss}ner, and Dai]{yeshwanthliu2023scannetpp}
Chandan Yeshwanth, Yueh-Cheng Liu, Matthias Nie{\ss}ner, and Angela Dai.
\newblock Scannet++: A high-fidelity dataset of 3d indoor scenes.
\newblock In \emph{Proceedings of the International Conference on Computer Vision ({ICCV})}, 2023.

\bibitem[Yin et~al.(2024)Yin, Liu, Xiao, Cohen-Or, Huang, and Chen]{yin2023sai3d}
Yingda Yin, Yuzheng Liu, Yang Xiao, Daniel Cohen-Or, Jingwei Huang, and Baoquan Chen.
\newblock Sai3d: Segment any instance in 3d scenes.
\newblock In \emph{Proceedings of the IEEE/CVF Conference on Computer Vision and Pattern Recognition (CVPR)}, 2024.

\bibitem[You et~al.(2023)You, Zhang, Gan, Du, Zhang, Wang, Cao, Chang, and Yang]{you2023ferret}
Haoxuan You, Haotian Zhang, Zhe Gan, Xianzhi Du, Bowen Zhang, Zirui Wang, Liangliang Cao, Shih-Fu Chang, and Yinfei Yang.
\newblock Ferret: Refer and ground anything anywhere at any granularity.
\newblock \emph{arXiv preprint arXiv:2310.07704}, 2023.

\bibitem[Yu et~al.(2023)Yu, Shen, and Chen]{yu2023towards}
Qihang Yu, Xiaohui Shen, and Liang-Chieh Chen.
\newblock Towards open-ended visual recognition with large language model.
\newblock In \emph{arxiv: 2311.08400}, 2023.

\bibitem[Yuan et~al.(2023)Yuan, Li, Liu, Tang, Luo, Qin, Zhang, and Zhu]{Osprey}
Yuqian Yuan, Wentong Li, Jian Liu, Dongqi Tang, Xinjie Luo, Chi Qin, Lei Zhang, and Jianke Zhu.
\newblock Osprey: Pixel understanding with visual instruction tuning, 2023.

\bibitem[Yue et~al.(2024)Yue, Chen, Geiping, Li, Goldstein, and Lim]{nxtp}
Kaiyu Yue, Bor-Chun Chen, Jonas Geiping, Hengduo Li, Tom Goldstein, and Ser-Nam Lim.
\newblock {Object Recognition as Next Token Prediction}.
\newblock In \emph{Computer Vision and Pattern Recognition Conference (CVPR)}, 2024.

\bibitem[Zhou et~al.(2017)Zhou, Zhao, Puig, Fidler, Barriuso, and Torralba]{ade}
Bolei Zhou, Hang Zhao, Xavier Puig, Sanja Fidler, Adela Barriuso, and Antonio Torralba.
\newblock Scene parsing through ade20k dataset.
\newblock In \emph{Proceedings of the IEEE conference on computer vision and pattern recognition}, pages 633--641, 2017.

\bibitem[Zhou et~al.(2022)Zhou, Girdhar, Joulin, Kr{\"a}henb{\"u}hl, and Misra]{detic}
Xingyi Zhou, Rohit Girdhar, Armand Joulin, Philipp Kr{\"a}henb{\"u}hl, and Ishan Misra.
\newblock Detecting twenty-thousand classes using image-level supervision.
\newblock In \emph{ECCV}, 2022.

\bibitem[Ziyu et~al.(2023)Ziyu, Xiaojian, Yixin, Zhidong, Siyuan, and Qing]{3dvista}
Zhu Ziyu, Ma Xiaojian, Chen Yixin, Deng Zhidong, Huang Siyuan, and Li Qing.
\newblock 3d-vista: Pre-trained transformer for 3d vision and text alignment.
\newblock In \emph{ICCV}, 2023.

\end{thebibliography}
}

\end{document}